\documentclass{amia}
\usepackage{booktabs}
\usepackage{multirow}
\usepackage{amsmath}
\usepackage{hyperref}
\usepackage{float}
\usepackage{enumitem}
\setlist{nosep,leftmargin=1.2em}

\begin{document}

\title{Beyond the Blood Draw: Explainable Machine Learning for Non-Invasive Dysglycemia Risk Screening}

\author{Black Sun, BS$^{1,*}$, Chenyi Zhang, BS$^{2,*}$, Kaiyi Ji, PhD$^{2}$, Xi Lu, PhD$^{2}$}

\institutes{
    $^1$Department of Computer Science, Aarhus University, Aarhus, Denmark\\
    $^2$University at Buffalo, SUNY, Buffalo, New York, USA
}

\maketitle
\begin{center}
\vspace{-15pt}
{\small $^*$Equal contribution.}
\end{center}
\vspace{-6pt}



\section*{Abstract}
\textit{Dysglycemia, encompassing both prediabetes and diabetes, affects huge numbers of adults worldwide, yet many of them remain undiagnosed. We developed and validated machine-learning (ML) models for non-invasive dysglycemia risk screening that require no laboratory tests. Pooling data from the National Health and Nutrition Examination Survey (NHANES) 2017--2023 (n=14,352), we trained six ML models with stratified 5-fold cross-validation and compared them with two established clinical risk scores. LightGBM achieved the highest area under the receiver operating characteristic curve (AUC=0.820, 95\% CI: 0.806--0.835), outperforming the Finnish Diabetes Risk Score (0.745) and American Diabetes Association Risk Test (0.783). SHAP analysis identified age, race/ethnicity, and waist-to-height ratio as the most influential predictors. Subgroup analyses confirmed consistent performance across demographic strata (AUC: 0.735--0.832). These results demonstrate the feasibility of explainable, laboratory-free dysglycemia screening for deployment in community settings and self-tracking health applications.}

\section*{Introduction}


Dysglycemia, a clinical spectrum encompassing both prediabetes and diabetes, is defined by chronic abnormalities in blood glucose levels. According to the American Diabetes Association (ADA) criteria, dysglycemia is primarily diagnosed using glycated hemoglobin (HbA1c), which reflects average glycemia over the preceding 2-3 months.\cite{ada2026diagnosis} A normal glycemic state is defined as HbA1c $< 5.7\%$, while prediabetes (HbA1c $5.7\%$--$6.4\%$) and diabetes (HbA1c $\geq 6.5\%$) represent progressive stages of metabolic impairment \cite{apa2023}. Dysglycemia typically begins at the prediabetes threshold (HbA1c $\geq 5.7\%$), which is a critical window where intervention can most effectively prevent progression to diabetes and its associated complications, such as retinopathy and cardiovascular disease \cite{tabak2012}. Globally, the public health burden caused by diabetes is immense. The International Diabetes Federation estimates that 589 million adults aged 20-79 currently live with diabetes, with projections reaching 853 million by 2050.\cite{idf2025} In the United States alone, 38.4 million people have diabetes, and 97.6 million have prediabetes.\cite{cdc2022} Alarmingly, 43\% of affected individuals remain undiagnosed,\cite{idf2025} missing the window for early interventions.\cite{tabak2012}

Current diagnostic approaches for dysglycemia are inherently \textbf{invasive}. According to ADA guidelines, a definitive diagnosis requires laboratory-based blood tests, most notably the glycated hemoglobin (HbA1c) test, where a level of $\geq$5.7\% indicates prediabetes and $\geq$6.5\% confirms diabetes.\cite{ada2026diagnosis} These procedures require venipuncture and laboratory processing, creating substantial barriers to frequent or large-scale screening in community-based and resource-limited settings. These \textit{invasive} procedures create substantial barriers to screening in resource-limited settings, including community pharmacies, workplace wellness programs, and underserved communities where phlebotomy infrastructure is unavailable. In contrast, \textbf{non-invasive} alternatives, questionnaire-based risk scores such as the Finnish Diabetes Risk Score (FINDRISC)\cite{lindstrom2003} and the American Diabetes Association (ADA) Risk Test\cite{bang2009}, require no blood draw but rely on linear additive scoring, which cannot capture the complex nonlinear interactions inherent in metabolic risk. Validation studies of these \textit{non-invasive} scores have shown variable performance (AUC 0.65--0.80).\cite{makrilakis2011,kengne2014} 

Machine learning (ML) methods can model nonlinear feature interactions and complex decision boundaries that linear scoring systems cannot represent. Recent studies have applied various ML algorithms to dysglycemia, especially diabetes prediction, with promising results.\cite{zou2018,kopitar2020,deberneh2021,zhang2020,liu2023machine} However, several methodological limitations persist: many studies include \textit{invasive} laboratory features (e.g., fasting glucose, lipid panels) in their predictor sets while claiming non-invasive screening;\cite{rafie2025leveraging,lee2015identification} most rely on random train-test splits without rigorous cross-validation; and few compare ML models head-to-head with established clinical risk scores. Additionally, model explainability and algorithmic fairness across demographic subgroups are rarely assessed despite their importance for clinical adoption.\cite{rudin2019,obermeyer2019} 
This study makes three primary contributions:
\begin{enumerate}
    \item We develop and validate ML models using a \textit{\textbf{strictly non-invasive}} feature set (self-report, basic anthropometry, and home blood pressure measurements) on recent NHANES data (2017--2023), using 5-fold stratified cross-validation and a systematic comparison against FINDRISC and the ADA Risk Test.
    \item We apply SHapley Additive exPlanations (SHAP)\cite{lundberg2017} to assess global and individual-level model interpretability, thereby enabling clinically transparent predictions.
    \item We evaluate algorithmic fairness across age, sex, race/ethnicity, and BMI subgroups and discuss the potential for deploying such models in self-tracking health-monitoring systems.
\end{enumerate}

\section*{Methods}

\begin{figure}[h]
  \centering
  \includegraphics[width=0.7\textwidth]{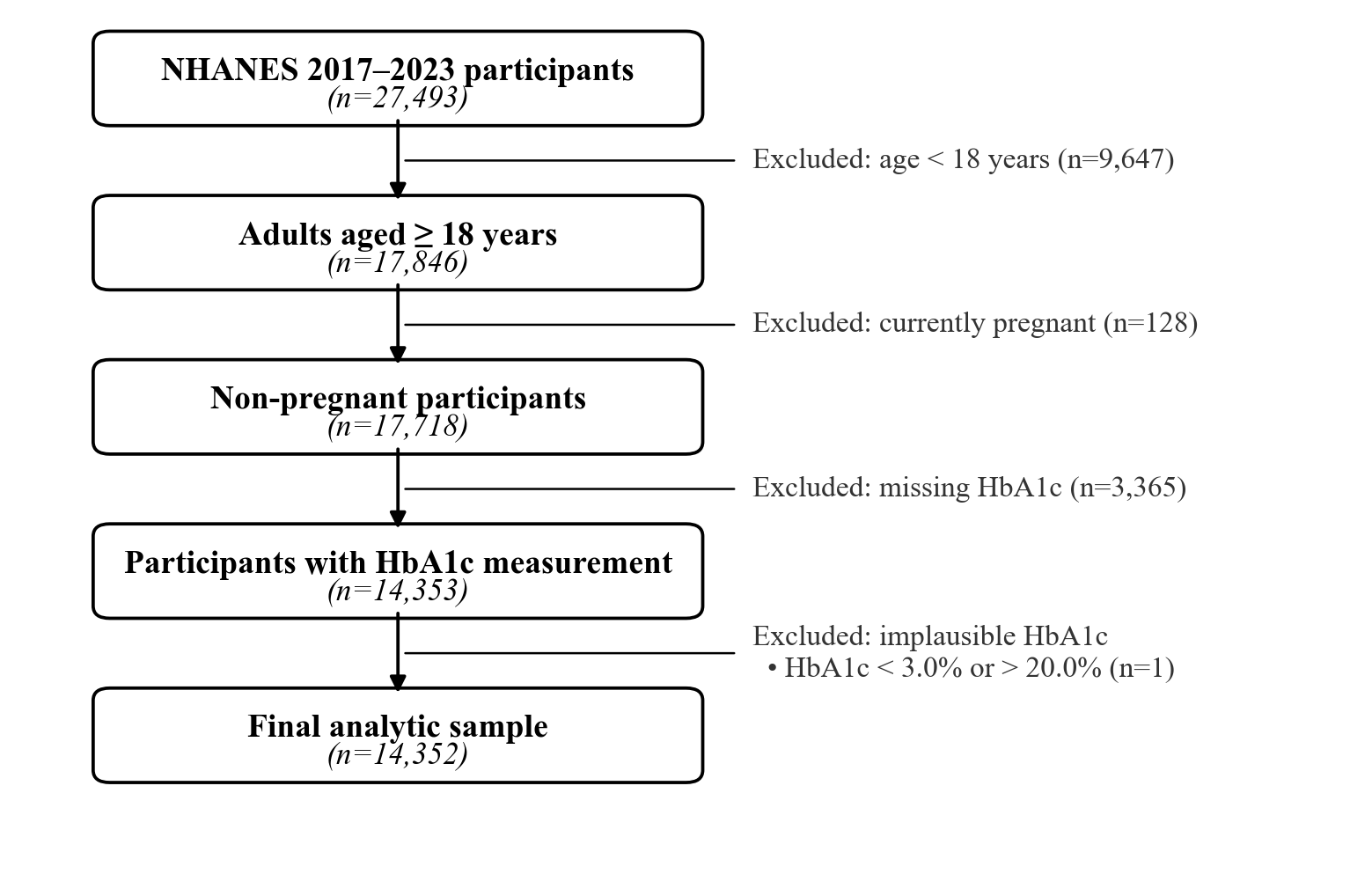}
  \caption{Participant selection flowchart from 27,493 NHANES 2017--2023 participants to the final analytic sample of 14,352 adults.}
  \label{fig:consort}
\end{figure}

\textbf{Data Source and Study Population.}
Data were obtained from the National Health and Nutrition Examination Survey (NHANES), conducted by the National Center for Health Statistics (NCHS).\cite{nhanes2024} NHANES uses a stratified, multistage probability sampling design to produce nationally representative health estimates for the U.S. civilian noninstitutionalized population. We pooled two recent cycles: the 2017--March 2020 pre-pandemic cycle and the August 2021--August 2023 cycle. Participants were included if they were aged $\geq$18 years, not pregnant, and had a valid HbA1c measurement within 3.0--20.0\%. These criteria yielded 14,352 eligible adults (Figure~\ref{fig:consort}). The pooled sample was randomly split 80:20 into training (n=11,481) and test (n=2,871) sets, stratified by outcome.

\textbf{Outcome Definition.}
The primary outcome was binary dysglycemia, defined as HbA1c $\geq$5.7\% or self-reported physician-diagnosed diabetes (DIQ010 = 1), consistent with ADA criteria.\cite{apa2023} HbA1c does not require fasting, is collected on all examined NHANES participants, and provides a stable measure of average glycemia over the preceding 2--3 months.

\textbf{Non-Invasive Feature Set.}
We defined ``non-invasive'' strictly as features obtainable through (1) self-report via brief questionnaire, (2) simple anthropometry (scale and tape measure), and (3) a consumer-grade blood pressure monitor. No blood draws, urine samples, or laboratory equipment are required. The 29 features span six domains:

\begin{itemize}
    \item \textbf{Demographics (5):} age, sex, race/ethnicity, education, and income-to-poverty ratio.
    \item \textbf{Anthropometry (4):} BMI, waist circumference, weight, and height.
    \item \textbf{Hemodynamics (2):} systolic and diastolic blood pressure. 
    \item \textbf{Medical/family history (4):} family diabetes history, hypertension diagnosis, antihypertensive medication use, and high cholesterol.
    \item \textbf{Lifestyle (10):} smoking status, alcohol frequency and quantity, occupational and recreational physical activity, sedentary time, sleep duration, and sleep disturbance.
    \item \textbf{Derived (4):} waist-to-height ratio, 10-year weight change percentage, PHQ-9 depression score, and gestational diabetes history.
\end{itemize}
Importantly, all 29 predictors are either stable demographic attributes, self-reported historical information, or measurements collected during the same NHANES examination visit as the HbA1c assay, but independently of it. None of the predictors is derived from or downstream of the HbA1c value, thereby ruling out label leakage from the outcome to the feature set.

\textbf{Missing Data Handling.}
Missing data were handled using multivariate imputation by chained equations (MICE).\cite{buuren2011mice} The imputer was fitted on the training set (10 iterations) and the fitted model was applied to the test set.

\textbf{Prediction Models.}
Eight models were compared: (i) FINDRISC,\cite{lindstrom2003} (ii) ADA Risk Test,\cite{bang2009} (iii) $L_2$-regularized logistic regression, (iv) random forest, (v) XGBoost,\cite{chen2016xgboost} (vi) LightGBM,\cite{ke2017lightgbm} (vii) SVM with RBF kernel, and (viii) multilayer perceptron (MLP). Hyperparameters for ML models were tuned via randomized search with 5-fold stratified cross-validation to optimize AUC.

\textbf{Evaluation.}
The primary metric was AUC. Secondary metrics included area under the precision-recall curve (AUPRC), sensitivity and specificity at the Youden-optimal threshold ($J = \text{sensitivity} + \text{specificity} - 1$), F1-score, and Brier score. 95\% confidence intervals for AUC were estimated via 1,000-iteration bootstrap resampling. Model calibration and clinical utility were evaluated using calibration curves and decision curve analysis (DCA).\cite{vickers2006dca} We note that formal pairwise statistical tests (e.g., DeLong test) for inter-model AUC differences were not conducted; therefore, model comparisons are based on point estimates and confidence interval overlap rather than hypothesis testing.

\textbf{Explainability and Fairness.}
We applied SHAP (SHapley Additive exPlanations)\cite{lundberg2017} using TreeExplainer to compute global feature importance, beeswarm summary plots, and individual waterfall plots. Algorithmic fairness was assessed by computing AUC with bootstrap 95\% CIs across subgroups defined by age, sex, race/ethnicity, and BMI.

\section*{Results}

\textbf{Study Population.}
From 27,493 NHANES 2017--2023 participants, sequential application of inclusion criteria excluded 9,647 individuals aged $<$18 years, 128 currently pregnant women, 3,365 participants without an HbA1c measurement, and 1 participant with an implausible HbA1c value, yielding a final analytic sample of 14,352 adults (Figure~\ref{fig:consort}). The training set comprised 11,481 individuals and the test set 2,871, with dysglycemia prevalence of 42.5\% in both sets, confirming successful stratified randomization. Table~\ref{tab:baseline} presents baseline characteristics by glycemic status.

As expected, participants with dysglycemia were substantially older (mean age 56.3 and 59.8 years for prediabetes and diabetes, respectively, versus 42.7 years for normoglycemic individuals), had higher BMI (31.0 and 33.0 vs.\ 27.8~kg/m$^2$), and exhibited greater central adiposity as measured by waist circumference (105.0 and 110.8 vs.\ 95.6~cm). The prevalence of comorbid conditions showed a clear dose-response relationship across glycemic strata: family diabetes history increased from 21.5\% (normal) to 31.4\% (prediabetes) to 51.0\% (diabetes), and hypertension rose from 23.2\% to 43.0\% to 65.0\%. These graded associations provide face validity for the selected non-invasive features and suggest that meaningful discriminative information is available without laboratory testing.

\vspace{6pt}
\begin{table}[h]
\caption{Baseline characteristics by glycemic status. Continuous: mean (SD); categorical: n (\%).}
\label{tab:baseline}
\begin{center}
\small
\begin{tabular}{lccc}
\toprule
\textbf{Characteristic} & \textbf{Normal} & \textbf{Prediabetes} & \textbf{Diabetes} \\
 & (n=8,256) & (n=3,579) & (n=2,517) \\
\midrule
Age, years & 42.7 (17.7) & 56.3 (16.7) & 59.8 (14.0) \\
BMI, kg/m$^2$ & 27.8 (6.6) & 31.0 (7.3) & 33.0 (7.9) \\
Waist circumference, cm & 95.6 (16.0) & 105.0 (16.1) & 110.8 (16.7) \\
Systolic BP, mmHg & 121.4 (16.7) & 128.8 (18.5) & 131.6 (19.9) \\
Male sex, n (\%) & 3,996 (48.4) & 1,754 (49.0) & 1,348 (53.6) \\
Family diabetes, n (\%) & 1,776 (21.5) & 1,124 (31.4) & 1,284 (51.0) \\
Hypertension, n (\%) & 1,918 (23.2) & 1,540 (43.0) & 1,636 (65.0) \\
\bottomrule
\end{tabular}
\end{center}
\end{table}

\begin{figure}[h]
  \centering
  \includegraphics[width=0.5\textwidth]{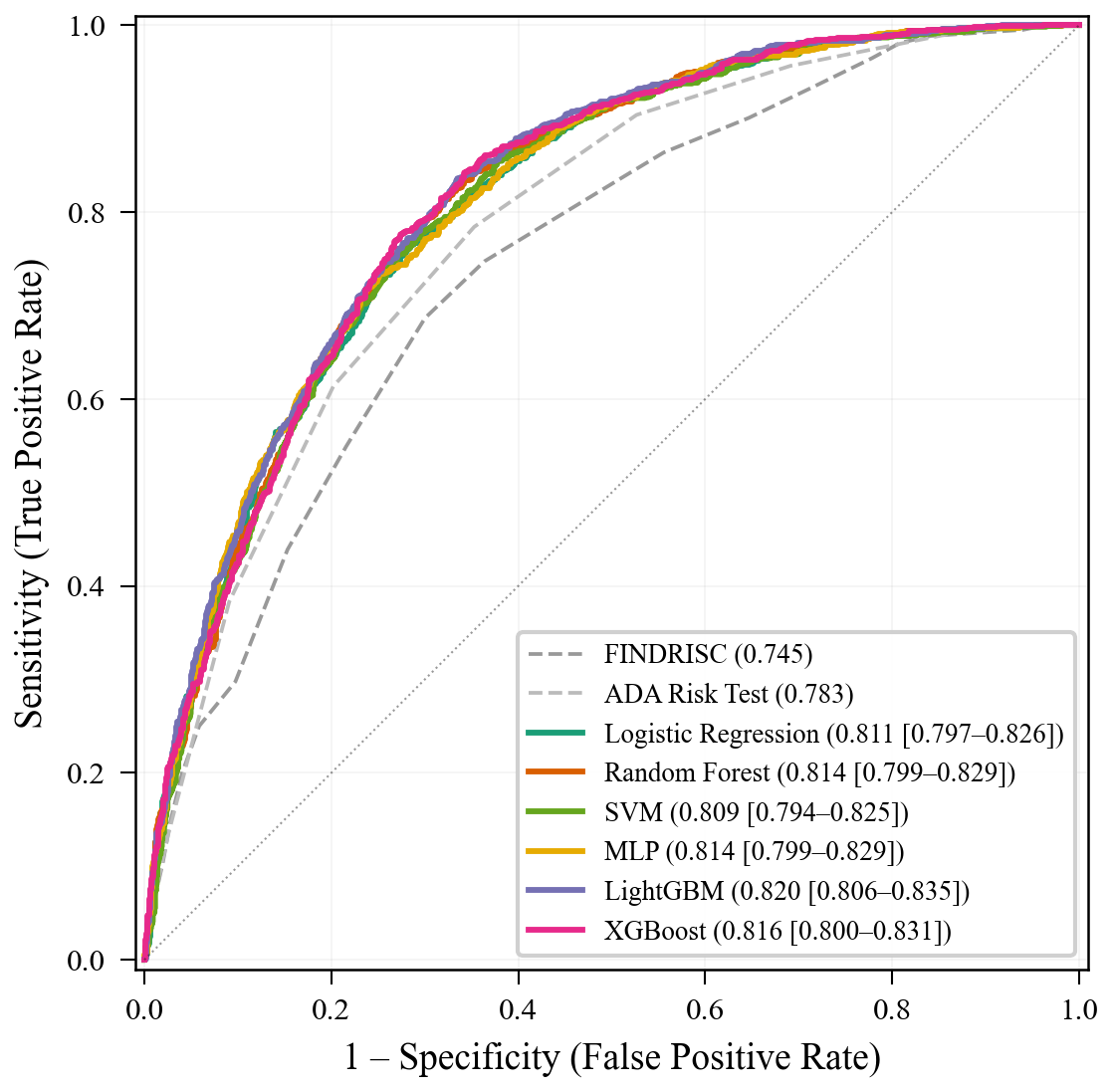}
  \caption{ROC curves for all eight models on the test set. Gradient-boosted tree models (LightGBM, XGBoost) achieve the highest discrimination, with all ML models (solid) outperforming traditional risk scores (dashed).}
  \label{fig:roc}
\end{figure}

\textbf{Discriminative Performance.}
Model performance on the held-out test set (n=2,871) is summarized in Table~\ref{tab:performance} and Figure~\ref{fig:roc}. Several key findings emerged.

\textit{Overall ranking.} LightGBM achieved the numerically highest AUC (0.820, 95\% CI: 0.806--0.835), followed by XGBoost (0.816, 0.800--0.831), MLP (0.814, 0.799--0.829), Random Forest (0.814, 0.799--0.829), and Logistic Regression (0.811, 0.797--0.826). SVM with RBF kernel had the lowest AUC among ML models (0.809, 0.794--0.825). Notably, the 95\% confidence intervals of all six ML models overlapped substantially, and the total AUC spread was only 0.011 (0.809--0.820), suggesting that the performance differences among ML models may not be statistically significant. This narrow spread indicates a discriminative ceiling of approximately 0.82 for this non-invasive feature set, implying that the performance bottleneck lies in the features' information content rather than in model expressiveness.

\textit{ML versus traditional risk scores.} All six ML models significantly outperformed both established clinical risk scores. LightGBM improved upon FINDRISC by 7.5 AUC percentage points (0.820 vs.\ 0.745) and upon the ADA Risk Test by 3.7 points (0.820 vs.\ 0.783). The ADA Risk Test (AUC = 0.783) outperformed FINDRISC (0.745) among traditional scores, likely because it incorporates a broader set of risk factors and was developed for the U.S. population. Even Logistic Regression (0.811), the simplest ML model, outperformed both traditional scores, suggesting that the improvement derives not only from model complexity but also from the richer feature set (29 variables versus 7--8 in FINDRISC/ADA).

\textit{Sensitivity-specificity trade-offs.} An important distinction among models emerges when examining sensitivity and specificity at the Youden-optimal threshold. LightGBM achieved the most sensitivity-oriented operating point (Sens = 0.837, Spec = 0.664), prioritizing detection of at-risk individuals at the cost of more false positives. XGBoost struck a more balanced trade-off (Sens = 0.776, Spec = 0.726), while MLP operated at the most specificity-dominant point (Sens = 0.734, Spec = 0.748). For a screening application where the primary goal is to identify individuals who should receive confirmatory HbA1c testing, higher sensitivity is clinically preferred because the cost of missing a true case (delayed diagnosis, progressive complications) far exceeds the cost of a false positive (an unnecessary but harmless blood test). From this perspective, LightGBM's high-sensitivity operating point is particularly well-suited to population-level screening.

\textit{AUPRC and class imbalance.} With 42.5\% dysglycemia prevalence, class imbalance is moderate. The AUPRC values (ranging from 0.652 for FINDRISC to 0.743 for LightGBM) closely tracked AUC rankings, confirming that performance differences are robust across both threshold-free metrics. The consistent LightGBM advantage in both AUC and AUPRC reinforces its selection as the preferred model.

\vspace{6pt}
\begin{table}[h]
\caption{Model performance on the test set (n=2,871). Sensitivity and specificity at Youden-optimal threshold. Bold indicates the numerically highest AUC among ML models; confidence intervals overlapped across all ML models.}
\label{tab:performance}
\begin{center}
\footnotesize
\begin{tabular}{lcccccc}
\toprule
\textbf{Model} & \textbf{AUC [95\% CI]} & \textbf{AUPRC} & \textbf{Sens.} & \textbf{Spec.} & \textbf{F1} & \textbf{Brier} \\
\midrule
FINDRISC & 0.745 [.728--.762] & 0.652 & 0.687 & 0.700 & 0.597 & 0.206 \\
ADA Risk Test & 0.783 [.767--.799] & 0.677 & 0.784 & 0.647 & 0.693 & 0.189 \\
\midrule
Logistic Reg. & 0.811 [.797--.826] & 0.731 & 0.764 & 0.716 & 0.710 & 0.178 \\
Random Forest & 0.814 [.799--.829] & 0.733 & 0.813 & 0.680 & 0.713 & 0.176 \\
XGBoost & 0.816 [.800--.831] & 0.739 & 0.776 & 0.726 & 0.703 & 0.173 \\
\textbf{LightGBM} & \textbf{0.820} [.806--.835] & \textbf{0.743} & 0.837 & 0.664 & 0.721 & 0.175 \\
SVM (RBF) & 0.809 [.794--.825] & 0.722 & 0.753 & 0.729 & 0.701 & 0.176 \\
MLP & 0.814 [.799--.829] & 0.737 & 0.734 & 0.748 & 0.694 & 0.174 \\
\bottomrule
\end{tabular}
\end{center}
\end{table}

\begin{figure}[H]
  \centering
  \includegraphics[width=0.9\textwidth]{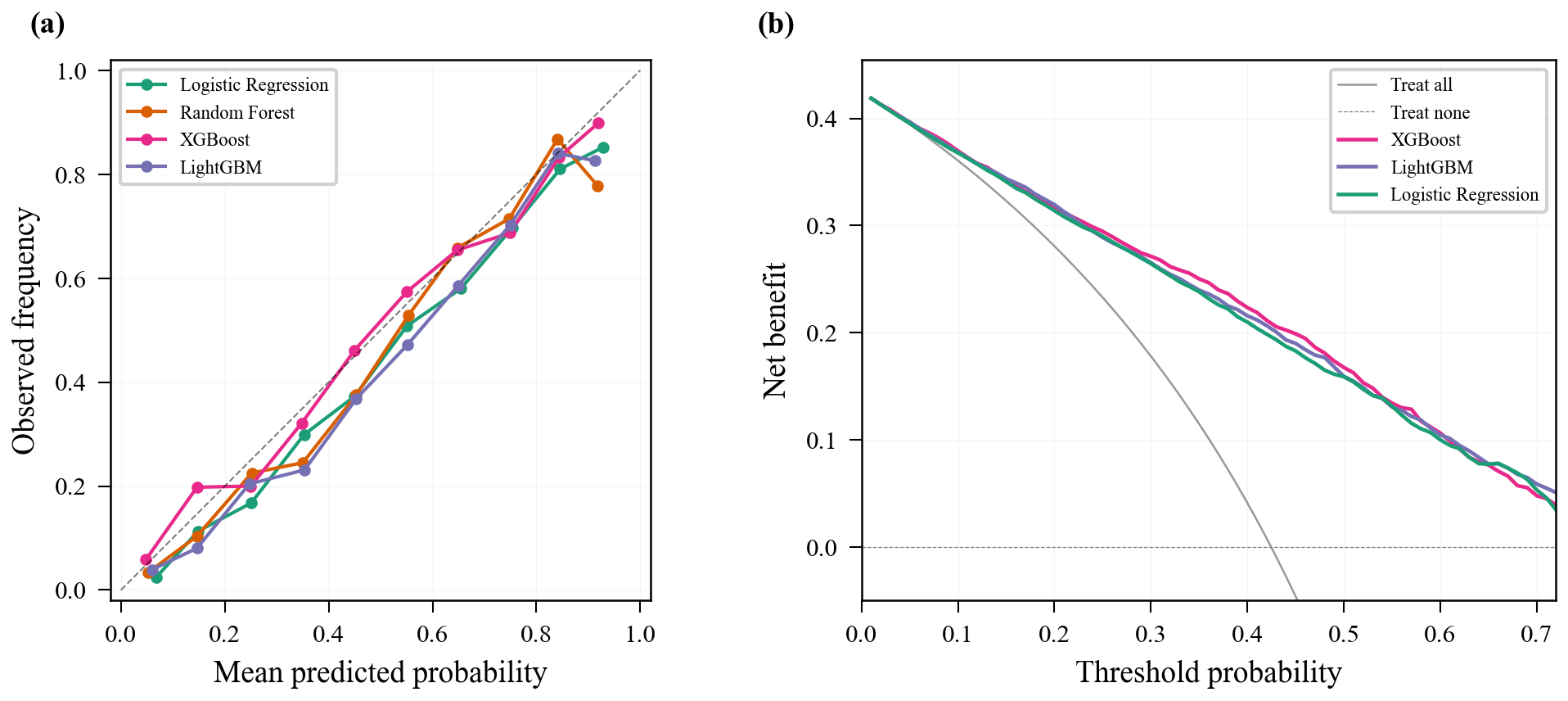}
  \caption{(a) Calibration curves comparing predicted probabilities with observed frequencies. (b) Decision curve analysis comparing net benefit across threshold probabilities.}
  \label{fig:cal_dca}
\end{figure}

\textbf{Calibration and Clinical Utility.}
Beyond discrimination, a screening model must produce well-calibrated probability estimates so that predicted risks correspond to actual observed frequencies. Calibration and clinical utility are presented in Figure~\ref{fig:cal_dca}.
\textit{Calibration (panel a).} LightGBM and XGBoost predictions closely followed the diagonal line of perfect calibration across the full range of predicted probabilities, indicating that their risk estimates can be meaningfully communicated to patients and clinicians. Random Forest showed mild overestimation in the mid-range (predicted probabilities of 0.3--0.5), while Logistic Regression exhibited slight overestimation at higher predicted probabilities ($>$0.6). Good calibration is particularly important for self-tracking applications, where users must trust that a reported ``35\% dysglycemia risk'' reflects a genuine one-in-three chance of having the condition, rather than an arbitrary score.
\textit{Decision curve analysis (panel b).} DCA evaluates whether using the model to guide screening decisions yields better outcomes than the default strategies of screening everyone (``treat all'') or screening no one (``treat none''). LightGBM provided the highest net clinical benefit across a clinically relevant range of threshold probabilities (approximately 10\%--60\%). At a threshold of 20\% (a plausible screening cutoff), LightGBM delivered substantially greater net benefit than both ``treat all'' and competing models. The net benefit advantage was most pronounced in the 15\%--45\% threshold range, which encompasses the range of risk thresholds where clinical decision-making is most uncertain. Below 10\%, all strategies converge toward ``treat all,'' and above 60\%, all converge toward ``treat none,'' consistent with theoretical expectations.

\begin{figure}[h]
  \centering
  \includegraphics[width=0.8\textwidth]{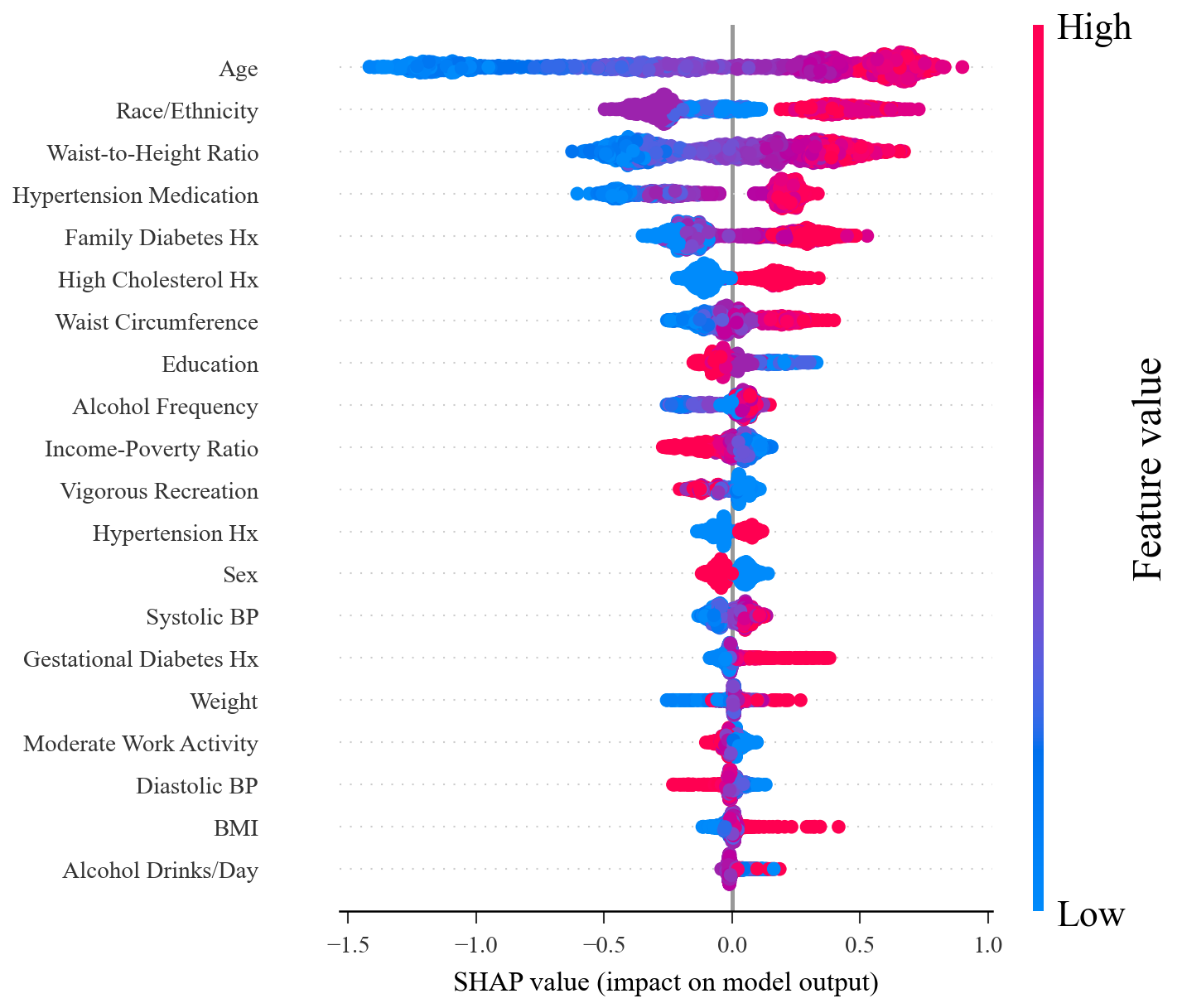}
  \caption{SHAP beeswarm plot for the LightGBM model. Each point represents one observation; horizontal position indicates SHAP value; color encodes feature value (red = high, blue = low).}
  \label{fig:shap}
\end{figure}

\textbf{Feature Importance and Explainability.}
SHAP analysis of LightGBM (Figure~\ref{fig:shap}) revealed age as the most influential predictor (mean $|$SHAP$|$ = 0.593), followed by race/ethnicity (0.298), waist-to-height ratio (0.288), antihypertensive medication use (0.258), and family diabetes history (0.210). High cholesterol history (0.136), waist circumference (0.093), education (0.071), and alcohol frequency (0.069) comprised the next tier of moderately important features.

These rankings are clinically coherent. Age reflects progressive beta-cell dysfunction and increasing dysglycemia over time. Waist-to-height ratio captures central adiposity, which is mechanistically linked to dysglycemia through visceral fat accumulation and systemic inflammation. Family diabetes history reflects shared genetic susceptibility and environmental exposures. The prominence of waist-to-height ratio over BMI supports growing evidence that central adiposity measures are superior for cardiometabolic risk stratification.\cite{tabak2012}

\begin{figure}[h]
  \centering
  \includegraphics[width=0.9\textwidth]{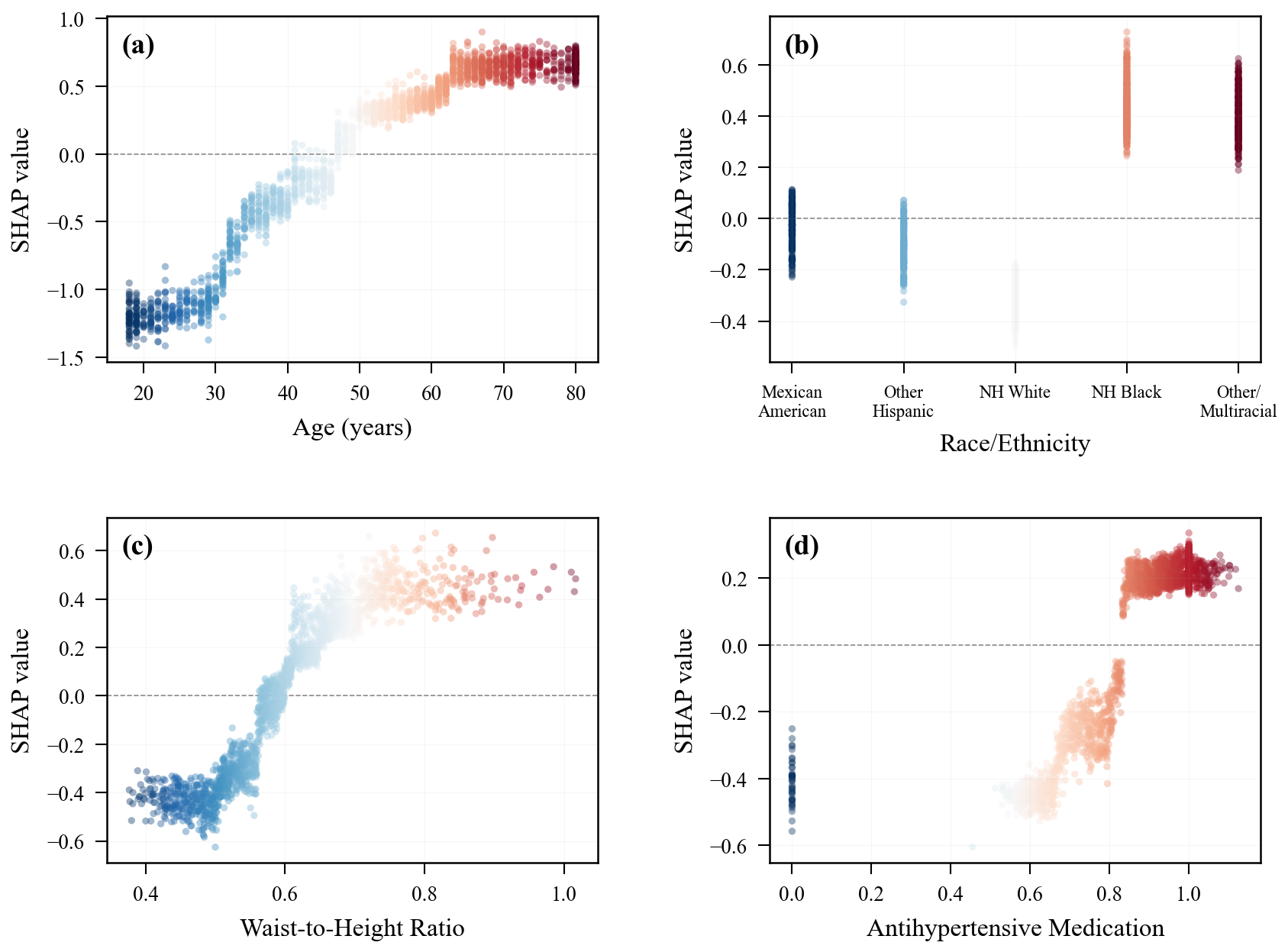}
  \caption{SHAP dependence plots for the four most influential features in the LightGBM model: (a) Age, (b) Race/Ethnicity, (c) Waist-to-Height Ratio, and (d) Antihypertensive Medication use. Each point represents one test-set observation. Horizontal axis shows the feature value; vertical axis shows the corresponding SHAP value (contribution to the log-odds prediction). Color encodes feature value from low (blue) to high (red). The dashed horizontal line indicates zero SHAP contribution.}
  \label{fig:shap_dep}
\end{figure}

\textbf{SHAP Dependence Analysis.}
SHAP dependence plots for the four most influential predictors (Figure~\ref{fig:shap_dep}) revealed clinically interpretable nonlinear patterns. Age (panel a) showed a monotonically increasing relationship, with a transition zone between ages 35 and 50 where SHAP values crossed zero---consistent with the ADA recommendation to screen all adults $\geq$35 years\cite{apa2023}---and a ceiling effect beyond age 55. Waist-to-height ratio (panel c) exhibited a sigmoidal curve: ratios below 0.50 were protective, a steep transition occurred between 0.55 and 0.70, and values above 0.75 showed diminishing marginal risk increases, suggesting saturation of visceral fat-mediated dysglycemia.\cite{tabak2012} The inflection near 0.50 aligns with established clinical cutoffs.\cite{browning2010} Race/ethnicity (panel b) showed discrete SHAP contributions that tracked known prevalence disparities,\cite{vyas2020} reinforcing that this variable captures social determinants rather than intrinsic biology. Antihypertensive medication use (panel d) produced a bimodal distribution, with medication users receiving positive SHAP contributions---likely reflecting comorbidity within the metabolic syndrome rather than a direct pharmacological effect on glycemia.

\begin{table}[h]
\caption{Subgroup-stratified AUC of LightGBM on the test set with 95\% bootstrap CIs.}
\label{tab:subgroup}
\begin{center}
\footnotesize
\begin{tabular}{llrcc}
\toprule
\textbf{Stratification} & \textbf{Category} & \textbf{N} & \textbf{Prevalence} & \textbf{AUC [95\% CI]} \\
\midrule
\multirow{3}{*}{Age group} & 18--39 & 878 & 14.5\% & 0.770 [0.729--0.808] \\
 & 40--59 & 877 & 47.1\% & 0.763 [0.734--0.794] \\
 & $\geq$60 & 1,116 & 60.8\% & 0.735 [0.706--0.765] \\
\midrule
\multirow{2}{*}{Sex} & Male & 1,393 & 45.6\% & 0.822 [0.798--0.842] \\
 & Female & 1,478 & 39.5\% & 0.821 [0.801--0.842] \\
\midrule
\multirow{4}{*}{Race/Ethnicity} & NH White & 1,314 & 36.3\% & 0.832 [0.810--0.853] \\
 & NH Black & 549 & 56.3\% & 0.764 [0.722--0.805] \\
 & Hispanic & 579 & 41.3\% & 0.814 [0.777--0.848] \\
 & Other & 429 & 45.2\% & 0.830 [0.789--0.867] \\
\midrule
\multirow{3}{*}{BMI category} & $<$25 & 743 & 22.9\% & 0.832 [0.795--0.863] \\
 & 25--29.9 & 937 & 43.8\% & 0.805 [0.776--0.832] \\
 & $\geq$30 & 1,191 & 53.7\% & 0.777 [0.752--0.805] \\
\bottomrule
\end{tabular}
\end{center}
\end{table}

\textbf{Subgroup and Fairness Analysis.}
Table~\ref{tab:subgroup} reports LightGBM AUC across demographic subgroups defined by age, sex, race/ethnicity, and BMI category. Thorough subgroup evaluation is essential for any screening tool intended for deployment across heterogeneous populations.

\begin{itemize}
    \item \textbf{\textit{Age.}} Model discrimination declined monotonically with age: AUC was 0.770 (95\% CI: 0.729--0.808) for adults aged 18--39, 0.763 (0.734--0.794) for ages 40--59, and 0.735 (0.706--0.765) for ages $\geq$60. This pattern is expected and well-understood: the baseline prevalence of dysglycemia rises sharply with age (14.5\%, 47.1\%, and 60.8\% across strata), which compresses the discriminative margin and reduces AUC as a statistical artifact. Importantly, the AUC remained above 0.73 even in the highest-prevalence group, indicating that the model retains meaningful discriminative capacity across the age spectrum. The high sensitivity of the model at its optimal threshold ensures that elderly individuals at risk are not systematically missed.
    \item \textbf{\textit{Sex.}} The model performed virtually identically between males (AUC = 0.822, 0.798--0.842) and females (0.821, 0.801--0.842), with completely overlapping confidence intervals. This parity is notable because diabetes risk factors differ somewhat by sex (e.g., a stronger obesity-diabetes risk association is observed in women, and visceral fat distribution differs between sexes \cite{kautzky2016sex}). The absence of a sex-based performance gap suggests that the model successfully captures sex-specific risk pathways without disadvantaging either group.
    \item \textbf{\textit{Race/Ethnicity.}} Performance varied across racial/ethnic groups: AUC was highest for Non-Hispanic White participants (0.832, 0.810--0.853) and the ``Other/Multiracial'' category (0.830, 0.789--0.867), followed by Hispanic participants (0.814, 0.777--0.848), and lowest for Non-Hispanic Black participants (0.764, 0.722--0.805). The 6.8 AUC percentage point gap between Non-Hispanic White and Non-Hispanic Black subgroups warrants attention. It is partly attributable to the substantially higher baseline prevalence in Non-Hispanic Black participants (56.3\% vs.\ 36.3\%), which compresses discrimination. However, it may also reflect differential accuracy of the HbA1c-based outcome definition across racial groups, as HbA1c levels are systematically higher in Non-Hispanic Black individuals at any given glucose level due to differences in hemoglobin glycation rates.\cite{apa2023} This highlights the need for external validation using multiple diagnostic criteria (e.g., fasting glucose and OGTT) in diverse populations.
    \item \textbf{\textit{BMI.}} Discrimination was highest for normal-weight individuals ($<$25 kg/m$^2$; AUC = 0.832, 0.795--0.863), intermediate for overweight (25--29.9; 0.805, 0.776--0.832), and lowest for obese ($\geq$30; 0.777, 0.752--0.805). The inverse relationship between BMI and AUC reflects the increasing prevalence of dysglycemia with higher BMI (22.9\%, 43.8\%, and 53.7\%), again compressing the discriminative range. Additionally, obese individuals have more heterogeneous metabolic profiles (metabolically healthy obesity vs.\ metabolically unhealthy obesity), making discrimination inherently more challenging. Despite this, the model maintained AUC $>$0.77 even in the obese subgroup, demonstrating clinically useful discrimination across the BMI spectrum.
\end{itemize}


\section*{Discussion}
This study demonstrates that ML models that use exclusively non-invasive features achieve clinically meaningful discrimination for dysglycemia screening (AUC range: 0.809--0.820), consistently outperforming two established clinical risk scores in contemporary NHANES data (2017--2023).

\textbf{Comparison with prior work.} Our results are consistent with the ML-based diabetes prediction literature (AUC 0.73--0.85),\cite{zou2018,kopitar2020,deberneh2021,zhang2020,liu2023machine} though direct comparisons are complicated by methodological heterogeneity. Critically, many prior studies inflate AUC through inclusion of laboratory features,\cite{rafie2025leveraging,lee2015identification} negating the practical advantage of non-invasive screening. Our work is distinguished by strict exclusion of all laboratory variables, head-to-head comparison with FINDRISC and the ADA Risk Test on the same dataset, SHAP-based explainability, and fairness evaluation across four demographic dimensions. A meta-analysis of 94 diabetes risk scores reported a median externally validated AUC of 0.74 (IQR: 0.71--0.78),\cite{kengne2014} placing our models in the upper range of reported performance without laboratory data.

\textbf{Clinical implications.} All six ML models performed within a narrow AUC band (0.809--0.820) with broadly overlapping confidence intervals, suggesting comparable discriminative ability and a performance ceiling of approximately 0.82 for this feature set. The logistic regression achieved an AUC of 0.811, indicating that most of the discriminative signal resides in main effects, while gradient-boosted trees may capture additional feature interactions (e.g., age $\times$ waist-to-height ratio) that account for their modest numerical advantage. From a deployment perspective, the choice among ML models may therefore be guided by secondary criteria such as calibration quality, sensitivity--specificity trade-off, and computational simplicity rather than AUC alone. For population-level screening, a high-sensitivity operating point (e.g., LightGBM: 83.7\% sensitivity, 66.4\% specificity) is clinically preferred: the cost of missing a true case --- delayed diagnosis and progressive complications --- far outweighs the cost of a false positive, which entails only a harmless confirmatory HbA1c test. In contrast, FINDRISC's lower sensitivity (0.687) would result in substantially more individuals remaining undiagnosed.

\textbf{Feature importance and clinical coherence.} SHAP rankings are clinically coherent, which is essential for clinician trust. Age dominated (mean $|$SHAP$|$ = 0.593), consistent with the exponential rise in dysglycemia incidence after age 40.\cite{apa2023} The prominence of waist-to-height ratio (0.288) over BMI supports evidence that central adiposity measures better capture cardiometabolic risk.\cite{browning2010} Notably, the SHAP dependence analysis revealed a risk transition near waist-to-height ratios of 0.50--0.55, aligning with established clinical cutoffs.\cite{browning2010} Race/ethnicity ranked second (0.298), reflecting well-documented prevalence disparities.\cite{vyas2020} However, race is a social construct serving as a proxy for environmental exposures, socioeconomic factors, and healthcare access rather than intrinsic biology. Future work should explore whether more granular social determinants of health (e.g., neighborhood deprivation indices) can maintain performance while reducing reliance on racial categorization.

\textbf{Fairness and equity.} Subgroup analysis showed consistent performance (AUC: 0.735--0.832) with no sex-based gap (male 0.822, female 0.821). The 6.8-point gap between Non-Hispanic White (0.832) and Non-Hispanic Black (0.764) participants is partly attributable to higher baseline prevalence compressing AUC, and partly to potential differential misclassification from HbA1c-based outcome definitions across racial groups.\cite{apa2023,obermeyer2019} Future work should evaluate equalized odds, predictive parity, and whether race-specific thresholds improve equitable sensitivity.

\textbf{Deployment in self-tracking applications.} All 29 features can be collected in 5--10 minutes without medical supervision, making the model well-suited for consumer-facing health applications and smartphone-based screening. Unlike clinical visits that require fasting and venipuncture, self-tracking enables continuous risk monitoring to capture temporal changes in modifiable factors. SHAP-based explanations can empower users by identifying the specific factors that drive their risk (e.g., age and waist-to-height ratio), thereby facilitating targeted preventive action. Future work should explore integrating wearable biosensor data (e.g., activity trackers, continuous glucose monitors) for real-time monitoring.

\textbf{Limitations.} NHANES is cross-sectional, precluding causal inference; prospective validation is needed to confirm that predicted risk translates to incident dysglycemia. HbA1c-based outcome definitions may misclassify individuals with hemoglobinopathies or altered erythrocyte lifespan,\cite{apa2023} particularly among Non-Hispanic Black participants, and self-reported variables are subject to recall and social desirability bias. The model lacks external validation outside NHANES, and generalizability to non-U.S. populations remains unknown. NHANES complex survey weights were not incorporated into training; however, because our objective is individual-level prediction rather than population-level inference, unweighted training is a defensible approach in predictive modeling, though the impact of sample weighting on calibration warrants future investigation. Formal pairwise statistical tests (e.g., DeLong test) were not performed to compare AUC values across models; given the narrow AUC spread (0.809--0.820) and broadly overlapping confidence intervals, the observed ranking among ML models should be interpreted as numerical rather than statistically established. Additionally, multicollinearity among the 29 predictors was not formally assessed. Although tree-based models are generally robust to correlated features, redundancy among anthropometric variables (e.g., BMI, waist circumference, weight, and waist-to-height ratio) may affect coefficient-based models such as logistic regression, and consolidating highly correlated features could improve model parsimony. Data from the 2021--2023 cycle may reflect pandemic-related behavioral shifts and selection bias. Finally, only two traditional risk scores were benchmarked; comparison with additional validated tools (e.g., Canadian Diabetes Risk Questionnaire, Leicester Risk Assessment Score) is warranted.

\textbf{Future directions.} Key priorities include: prospective clinical validation linking predicted risk to incident dysglycemia; external validation in non-NHANES cohorts (e.g., UK Biobank \cite{ukbiobank2026}, electronic health records); a randomized trial comparing ML-guided screening with standard-of-care approaches; and deployment in a smartphone application with iterative user experience testing.

\section*{Conclusion}
We developed and validated ML models for laboratory-free dysglycemia screening using 29 non-invasive features in 14,352 U.S. adults from NHANES 2017--2023. LightGBM achieved an AUC of 0.820, outperforming FINDRISC and the ADA Risk Test, with consistent performance across age, sex, racial/ethnic, and BMI subgroups. SHAP-based explainability identified clinically coherent predictors including age, waist-to-height ratio, and family diabetes history. The exclusively non-invasive nature of the model supports potential deployment in community screening programs and consumer-facing self-tracking health applications where laboratory diagnostics are unavailable.

\section*{Acknowledgments} 

This research used publicly available data from the National Health and Nutrition Examination Survey (NHANES), administered by the National Center for Health Statistics, CDC.

\makeatletter
\renewcommand{\@biblabel}[1]{\hfill #1.}
\makeatother

\bibliographystyle{vancouver}
\bibliography{amia}

\end{document}